\documentclass[runningheads]{llncs}
\usepackage[T1]{fontenc}
\usepackage{color}
\usepackage{placeins}
\usepackage{comment}
\usepackage{graphicx}
\usepackage{tabularx}
\usepackage{soul}
\usepackage{multirow}
\usepackage{multicol}
\usepackage{colortbl}
\usepackage{tikz}
\usepackage{amsmath}
\usepackage{euflag}
\makeatletter
\newcommand{\@final}{2}
\newcommand{\@todobase}[3]{\colorbox{#1}{\textbf{#2}}\hspace{5pt}\textcolor{#1}{#3}}
\newcommand{\@todo}[2][red]{\@todobase{#1}{ToDo:}{#2}}
\newcommand{\@todostar}[3][red]{\@todobase{#1}{#2}{#3}}
\newcommand{\todo}{\@ifstar{\@todostar}{\@todo}}
\newcommand{\@rvplus}[3]{
\if\@final0%
\@todobase{#1}{#2}{#3}%
\else%
#3
\fi%
}
\newcommand{\plus}[2]{\@rvplus{green}{#1}{#2}}
\newcommand{\@rvminus}[3]{
\if\@final0%
\@todobase{#1}{#2}{\sout{#3}}%
\fi%
}
\newcommand{\minus}[2]{\@rvminus{red}{#1}{#2}}
\newcommand{\anonymous}[2][\empty]{
\if\@final1%
#1%
\else%
#2%
\fi%
}
\makeatother
\graphicspath{{./images/}}
\usepackage{hyperref}

\begin{document}
\title{Evaluating the Reliability of Self-Explanations in Large Language Models}
%
%\titlerunning{Abbreviated paper title}
% If the paper title is too long for the running head, you can set
% an abbreviated paper title here
%
\author{Korbinian Randl\inst{1}\orcidID{0000-0002-7938-2747} \and
John Pavlopoulos\inst{1,2,3}\orcidID{0000-0001-9188-7425} \and
Aron Henriksson\inst{1}\orcidID{0000-0001-9731-1048} \and
Tony Lindgren\inst{1}\orcidID{0000-0001-7713-1381}}
\authorrunning{K. Randl et al.}
% First names are abbreviated in the running head.
% If there are more than two authors, 'et al.' is used.
%
\institute{Stockholm University, Department of Computer and Systems Sciences, Kista, SE-164 07, Sweden \email{\{korbinian.randl,ioannis,aronhen,tony\}@dsv.su.se} \and
Athens University of Economics and Business, Patission 76, Athens 104 34, Greece
\email{\{annis\}@aueb.gr} \and
Archimedes/Athena RC}
\maketitle              % typeset the header of the contribution
\begin{abstract}
This paper investigates the reliability of explanations generated by large language models~(LLMs) when prompted to explain their previous output. We evaluate two kinds of such self-explanations -- extractive and counterfactual -- using three state-of-the-art LLMs (2B to 8B parameters) on two different classification tasks (objective and subjective). 
Our findings reveal, that, while these self-explanations can correlate with human judgement, they do not fully and accurately follow the model's decision process, indicating a gap between perceived and actual model reasoning. 
We show that this gap can be bridged because prompting LLMs for counterfactual explanations can produce faithful, informative, and easy-to-verify results. These counterfactuals offer a promising alternative to traditional explainability methods (e.g., SHAP, LIME), provided that prompts are tailored to specific tasks and checked for validity. %Our results suggest that extractive self-explanations provide the most probable explanations based on training data, but can not generally be trusted to represent the inference process of the LLM. Conversely, counterfactual self-explanations, when carefully crafted, can enhance our understanding of whether LLMs have correctly understood assigned tasks. 

\keywords{Large Language Models \and Self-Explanations \and Counterfactuals.}
\end{abstract}
\section{Introduction}
In recent years, large language models~(LLMs) have made significant progress in natural language processing tasks, exhibiting impressive capabilities across various domains. Following their successes, these models have found their way into people's everyday lives, for example in the form of chatbots such as ChatGPT. In light of this great impact of the technology, and the increasing amounts of trust placed on it, a critical question remains: How reliable are the explanations these models provide for their own outputs and can they successfully explain their own reasoning processes? Understanding the internal reasoning of LLMs is crucial for building trust and transparency in their usage. This paper investigates the reliability of self-explanations generated by prompting LLMs to explain their previous outputs and provides the following contributions:
\begin{enumerate}
    \item We evaluate extractive self-explanations generated by three state-of-the-art LLMs (2B to 8B parameters) on two different classification tasks (objective and subjective), and show that, while these extracts may often seem intuitive for humans (as they show high correlation with human assessment) they are not guaranteed to fully and accurately describe the model's decision process.

    \item We show that the gap highlighted in our first contribution can be bridged. Specifically, our findings show that prompting the LLM for counterfactual explanations can create faithful explanations that can easily be validated by the model.

    \item We provide an analysis of counterfactual self-explanations created by LLMs and show that they can be highly faithful and similar to the original, but need to be individually checked for validity.
\end{enumerate}

\section{Related Work}

\subsubsection{Local Explainability for Transformers:}
In the scope of this work, we define LLMs as pre-trained text-to-text processing systems, based on the Transformer architecture~\cite{Vaswani2017_Transformer}. As such systems usually complete an input text by iteratively predicting the next token, we use the following notation throughout this paper:
\begin{equation}
    t_{n+1} = \mathrm{LLM}(t_0, ... , t_n)
\end{equation}
Specifically, LLMs consist of an embedding layer
$h^{(0)}_i=f^\mathrm{In}(t_i, i),$
computing the input embedding~$\textbf{h}^{(0)} = [h^{(0)}_0, ... , h^{(0)}_n],$
followed by $L$ transformer blocks~$\textbf{h}^{(l+1)} = f^{(l)}(\textbf{h}^{(l)}), 0 \le l < L,$
and a head~$t_{n+1}=f^\mathrm{out}(\textbf{h}^{(L)}).$
Each of the transformer blocks uses multi-head attention which we will explain in more detail later.

Modern Transformers can be divided into three sub-architectures:
encoder-only~\cite{Devlin2019_BERT,Liu_roberta}, encoder-decoder~\cite{Raffel2020_T5,Lewis2020_bart}, and decoder-only~\cite{gemmateam2024_gemma,llama3modelcard}.
%\textbf{(i)~encoder-only} type models, like BERT~\cite{Devlin2019_BERT} or RoBERTa~\cite{Liu_roberta}, employ only the encoder part of the original Transformer architecture, take text as an input, and map it to a latent space using only self-attention. They are usually extended with one or more task-specific output layers and fine-tuned on task-specific data.
%\textbf{(ii)~encoder-decoder} models follow the original design of Vaswani et al.~\cite{Vaswani2017_Transformer}, employing both self and cross-attention, and are intended for text-to-text tasks like machine translation. Current implementations are T5~\cite{Raffel2020_T5} and BART~\cite{Lewis2020_bart}.
%\textbf{(iii)~decoder-only} approaches, like Gemma~\cite{gemmateam2024_gemma}, Llama~\cite{Touvron2023_Llama2,llama3modelcard}, or the GPT family~\cite{Brown2020_FewShotLearning}, are largely similar in structure to encoder-only models but follow an iterative approach to completing an input prompt token by token. As they are trained to iteratively predict the next token, the input is aligned to the right of the context window (instead of left as in encoder-only), and the next token is predicted from the output corresponding to the last input token. 
%
Decoder-only LLMs have demonstrated good classification abilities, even without additional fine-tuning, using ``in-context learning'': By simply asking the LLM to classify a sample text provided in the input text or ``prompt'' along with a list of possible classes, LLMs can successfully solve many reasoning tasks.
This method called ``zero-shot prompting'' can be extended to ``few-shot prompting''~\cite{Brown2020_FewShotLearning} for more difficult tasks by additionally including a small number of labeled samples in the prompt. Recently, \textit{instruction-tuning} improves further the performance \cite{gemmateam2024_gemma,llama3modelcard}.

In this paper, we focus on local explainability. This means we want to explain specific predictions of the Transformer rather than explain how the model works in general.
Since the first publication in 2017, different methods for generating such explanations for the classification output of Transformers, and therefore LLMs, have been proposed. These are heavily dependent on the classification paradigm \cite{Zhao2024_XLLM}:
in general deep learning, which is often extendable to Transformers, the literature shows feature attribution approaches (e.g. relevance-propagation-based~\cite{Bach2015_LRP,Montavon2017_DeepTaylor} or gradient-based~\cite{Hechtlinger2016_InputGradient,Sundarajan2017_IntegratedGradients}). For traditional applications where the Transformer is fine-tuned to produce class probabilities via a task-specific output layer, we are aware of attention-based~\cite{Abnar2020_AttentionFlow} or mixed~\cite{Chefer2021_TransformerInterpretabilityBeyondAttentionVisualization,Liu2021_AttentionBasedExplanation} approaches. 
In prompting, specifically for instruction-tuned models, the literature mainly contains methods for generating textual explanations such as Chain-of-Thought~(CoT)~\cite{Wei2022_CoT,Zhao2024_XLLM}.
Independent of the applied paradigm, surrogate-based model-agnostic approaches, such as LIME~\cite{Ribeiro2016_LIME} and SHAP~\cite{Lundberg2017_SHAP}, can be found in literature~\cite{Li2023_UnderstandingICL,Zhao2024_XLLM}. Next, we introduce the most important types of explanations.

\subsubsection{Attention-Based Explanations}
leverage the Transformer's scaled dot product attention weights $A^{(l)}$, generated during the forward pass, to explain the impact of each input token to each output token. Given an input vector $\mathbf{h}^{(l)} \in \mathbb{R}^{n}$, the self-attention variant, which is applied in all Transformers used in this paper, is computed by feeding $\mathbf{h}^{(l)}$ through three linear layers, computing the vectors $\mathbf{q}^{(l)}$, $\mathbf{k}^{(l)}$, and $\mathbf{v}^{(l)}$, and then:
\begin{equation}
    A^{(l)} = \mathrm{softmax}\left(\frac{\mathbf{q}^{(l)} \cdot {\mathbf{k}^{(l)}}^T}{\sqrt{n}}\right)
\end{equation}
\begin{equation}
    \mathbf{h}'^{(l)} = A^{(l)} \cdot \mathbf{v}^{(l)}
\end{equation}
This makes $A^{(l)}$, with all elements $\in [0,1]$ a weight matrix connecting $\mathbf{h}^{(l)}$ and $\mathbf{h}'^{(l)}$.  
As Transformers produce one matrix $A$ per attention head (e.g. $12 \times 12$ heads in BERT$_\mathrm{base}$~\cite{Devlin2019_BERT}, $28 \times 16$ heads in Gemma-7B~\cite{gemmateam2024_gemma}) extracting meaningful explanations is not trivial: while naive approaches simply use the mean attention weights of the last layer, methods that follow the attention through the whole Transformer have been shown to outperform them~\cite{Abnar2020_AttentionFlow}. Attention-based explanations can be improved by combining them with gradient-based methods to estimate the importance of each head towards the prediction~\cite{Chefer2021_TransformerInterpretabilityBeyondAttentionVisualization}, the crucial step lies in connecting attention weights in the last attention layer to the output, as the residual connections within the Transformer keep input to output association stable over multiple layers~\cite{Liu2021_AttentionBasedExplanation}.
As there is a debate in the literature about whether attention weights can be used as explanations, we also employ gradient-based explanations.

\subsubsection{Gradient-Based Explanations}
create saliency maps of the input by computing the gradient~$\frac{\partial \mathrm{LLM}(\cdot)}{\partial h^{(0)}_i}, 0 \le i \le n$ of the $n+1^{th}$ output of the LLM with regard to a specific input embedding~$h^{(0)}_i$. In the simplest case, this gradient itself can be the explanation~\cite{Hechtlinger2016_InputGradient}, but the literature shows that computing the Hadamard product with the input improves on it~\cite{Shrikumar2017_GradIn}.
An often discussed problem of gradient-based approaches is the so-called saturation problem~\cite{Shrikumar2017_GradIn,Sundarajan2017_IntegratedGradients}: as neural networks minimize the absolute gradient during training, gradients of a well-fitted network will be close to zero. We argue, however, that the ambiguous nature of natural language prevents overfitting and therefore also, to a certain degree, gradient saturation of pre-trained multi-purpose LLMs. To support this theory, we provide statistics on the gradients for each of our experiments.

\subsubsection{Counterfactual Explanations}
are -- simply put -- versions of the model input that alter the model's output. 
A good counterfactual should fulfill at least the following two criteria~\cite{Verma2022_counterfactual}: \textbf{(i)~validity:}~the model output between the counterfactual and the original input should differ at inference time. \textbf{(ii)~similarity}~the changes made to the original to produce the counterfactual should be minimal: the more that is changed from the original, the less specific the counterfactual becomes, eventually making it irrelevant as a local explanation. However, defining a good distance measure for comparing two texts is non-trivial, and may well require the combination of both semantic and syntactic similarity. Therefore, this second point is sometimes overlooked in similar studies~\cite{madsen2024_selfexplanations}.

\vspace{-3pt}
\subsubsection{Rationale-Based Explanations}
can be described as textual excerpts or abstractions of the model input that contribute to the model's predictions~\cite{Gurrapu2023_Rationalization}. In contrast to feature attribution methods, these do not provide a measure for the importance of each token, but rather a text that describes the influences. While traditional methods rely on extraction or abstraction methods for generating those texts~\cite{Gurrapu2023_Rationalization}, LLMs can be prompted to provide explanations. CoT generates textual rationales at inference time and even boosts reasoning performance~\cite{Wei2022_CoT}, while additional prompting for \textit{self-explanations} has recently received increased attention in the research community:

Huang et al.~\cite{huang2023_CanLLMs} prompt ChatGPT to yield feature importance scores for a sentiment classification task based on 100 random texts taken from the Stanford Sentiment Treebank dataset~\cite{Socher2013_SST}. Then, they evaluate the faithfulness of the generated scores compared to LIME~\cite{Ribeiro2016_LIME} and occlusion~\cite{Li2017_occlusison}. They conclude that none of the three methods has a clear advantage over the others and that the methods often do not agree on feature importance. However, this ambiguity may be owed to their relatively small sample size. Furthermore, prompting an LLM to produce numerical importance scores for each token is not what these models are designed to produce.

Madsen et al.~\cite{madsen2024_selfexplanations} instead prompt for the most important words, counterfactuals, and redactions (i.e. asking the model to mask important tokens) for classification tasks on different datasets using Llama2~\cite{Touvron2023_Llama2}, Falcon~\cite{Almazrouei2023_Falcon}, and Mistral~\cite{Jiang2023_Mistral}. They conclude that faithfulness of explanations is highly dependent both on the choice of the LLM and on the data used. The authors, however, do not compare the extracted self-explanations to established explainability methods or human annotations.

\vspace{-3pt}
\paragraph{We extend} the previously discussed work by addressing the following questions:

\vspace{-3pt}
\begin{itemize}
    \item\textbf{RQ1:} Do LLM \textit{self-explanations} correlate well with human judgment?

    \item\textbf{RQ2:} Do LLM \textit{self-explanations} correlate well with internal model dynamics, represented by attention- and gradient-based explainability methods?
\end{itemize}

\section{Method}
To address our research questions, first we extract LLM self-explanations and explanations from model-specific explainability gradient/attention methods.

\subsection{Self-Explanations}

In all our experiments, we use zero-shot prompting using the chat-completion format. A typical chat can be seen in Table~\ref{tab:example_chat}: \textbf{a)}~As a first step, we prompt the LLM to perform the actual task: we study food hazard classification and sentiment classification . Following this, we generate two different types of self-explanations:
\textbf{b)~extractive self-explanations} are created by asking the LLM for the words/phrases (we ask the model to produce the same number of phrases as in the human ground truth) that were most important for its classification;
\textbf{c)~counterfactual self-explanations} are generated by asking the LLM to provide a version of the classified text, for which its decision would have been different and which has as few words as possible changed from the original text.
These two kinds of explanations are generated in separate chats to avoid cross-influencing the LLM's output.

\begin{table}[h]
    \centering

    \caption{Example Chat Sequence: Our approach is split into three steps: \textbf{a)} the actual task, \textbf{b)} extracting a posthoc self-explanation, and \textbf{c)} extracting a counterfactual. The latter two are performed in independent chat sequences. This example was produced using Llama~3-8B. The prompt is slightly adapted to fit different tasks and models.}
    \label{tab:example_chat}

    \resizebox{\textwidth}{!}{
        \renewcommand{\arraystretch}{1.5}
        \begin{tabular}{c l p{\linewidth}}

        \hline
        
        \multirow{2}{*}{\huge a)} &
        \textbf{user}: &
        \texttt{What is the sentiment of the following review?}
        
        \texttt{``Hints are made to the audience that this film could be a blast. Alas, these are only hints.''}
        
        \texttt{Assign one of the following labels: "negative" or "positive". Make sure to answer only with the label.} \\
        
        &
        \textbf{assistant}: &
        \texttt{Negative} \\

        \hline
        
        \multirow{2}{*}{\huge b)} &
        \textbf{user}: &
        \texttt{What is the most important phrase influencing your assessment? Provide only the phrase as a string.} \\
        
        &
        \textbf{assistant}: &
        \texttt{``Alas, these are only hints.''} \\

        \hline
        
        \multirow{2}{*}{\huge c)} &
        \textbf{user}: &
        \texttt{Provide a version of the review that would flip your assessment while changing as few words in the original review as possible. Make sure to answer with only the new version.} \\
        
        &
        \textbf{assistant}: &
        \texttt{``Hints are made to the audience that this film could be a blast. And indeed, these are more than just hints.''} \\

        \hline
        
        \end{tabular}
    }
\end{table}

\subsection{Model-Specific Explanations}
We compare the generated self-explanations to established explainability methods. In this paper, we do not consider explanation methods based on relevance propagation~\cite{Bach2015_LRP,Montavon2017_DeepTaylor}, as we are not aware of any implementations that sufficiently deal with the residual connections in the Transformer architecture. Instead, we focus on gradient- and attention-based methods, as these remain close to the internal information representation of the LLMs. Furthermore, we focus on \textit{decoder-only Transformers}, needed to generate self-explanations, which are largely similar to \textit{encoder-only Transformers} in architecture and therefore often allow for the direct application of attention-based explainability methods tested on BERT.
%This is not a limitation, as the top-performing LLMs at the time of writing employ the decoder-only architecture~\cite{gemmateam2024_gemma,Touvron2023_Llama2,llama3modelcard,Brown2020_FewShotLearning}.
In this work, we apply three simple analytic explanation methods:\\
\textbf{(i)~AGrad~\cite{Liu2021_AttentionBasedExplanation}} for an output~$t_{n+1}$ is the product:
\begin{equation}
   \frac{1}{H} \sum_{h=1}^{H} \frac{\partial\mathrm{LLM}(\cdot)}{\partial A^{(L)}_{h, j, k}} A^{(L)}_{h, j, k},
\end{equation}
where $A^{(L)}_{h, j, k}$ is the attention weight with indices $j,k$ in the $h^{th}$ head of the last layer $L$. As the authors of the method show, backpropagating the produced saliency to the input does not greatly improve faithfulness, because of the transformer's residual connections. Therefore, we omit this step.
\\\textbf{(ii)~Gradient times Input (GradIn)~\cite{Shrikumar2017_GradIn}} is a simple gradient-based method, computed by taking the product
\begin{equation}
    \frac{\partial \mathrm{LLM}(\cdot)}{\partial h^{(0)}_i} h^{(0)}_i,
\end{equation}
for an output token $t_{n+1}$ and an input embedding $h^{(0)}_i$.
\\\textbf{(iii)~Inverted Gradient (IGrad)} is our own approach to gradient-based counterfactual explanations, as we are not aware of existing methods in this domain. While GradIn approximates the impact of a change in the input towards the output of the network, IGrad takes a counterfactual approach by approximating the necessary change at the input in order to achieve a specific change of the output. Starting from the first-order Taylor approximation of an LLM
\begin{equation*}
    {\bf h}^{(L)} \approx \tilde{\bf h}^{(L)} +
    \left . \mathbb{J}_\mathrm{LLM}({\bf h}^{(0)}) \right | _{{\bf h}^{(0)}=\tilde{\bf h}^{(0)}} \cdot
    \left({\bf h}^{(0)} - \tilde{\bf h}^{(0)} \right),
\end{equation*}
we define the importance as the pseudo-inverse of the Jacobian~$\mathbb{J}_\mathrm{LLM}({\bf h}^{(0)})$ evaluated at the input embedding $\tilde{\bf h}^{(0)}$ and the corresponding output embedding $\tilde{\bf h}^{(L)}$. Therefore, the final explanation
\begin{equation}
    \tilde{\bf h}^{(0)} - {\bf h}^{(0)} \approx
    \left(\left . \mathbb{J}_\mathrm{LLM}({\bf h}^{(0)}) \right | _{{\bf h}^{(0)}=\tilde{\bf h}^{(0)}}\right)^{-1} \cdot
    \left(\tilde{\bf h}^{(L)} - {\bf h}^{(L)} \right)
\end{equation}
is a measure of how a specific change of the output maps to the LLM's input.

\section{Empirical Analysis}
In this section, 
%we describe an experimental analysis we undertook. 
we first describe the tasks we consider, then the evaluation metrics, and lastly the results per task. We perform our experiments with Google's \textit{Gemma~1.1 Instruct}~\cite{gemmateam2024_gemma} (\texttt{2B} and \texttt{7B} to assess model size impact) and Meta's \textit{Llama~3 Instruct}~\cite{llama3modelcard} (\texttt{8B}) from huggingface\footnote{\href{https://huggingface.co}{https://huggingface.co}} leveraging their chat formats.
%At the time of writing, those two models are among the best performing open-source LLMs.
All our experiments have been performed using 8 NVIDIA RTX A5500 graphics cards with 24GB of memory each. Our code is publicly available on GitHub\footnote{\href{https://github.com/k-randl/self-explaining_llms}{k-randl/self-explaining\_llms}}.

\subsection{Tasks and Data}
We assess the previously described methods on two different tasks:
\begin{enumerate}
    %\item \textbf{Entity extraction} on the \textit{Food Recall Incidents dataset}~\cite{Randl2024_FoodIncidents}. It contains the titles of official food recalls released by government and non-government organizations. We use expert annotations\footnote{Experts from \href{https://agroknow.com}{AgroKnow}}, identifying the supplier of the recalled product, to extract matching spans in the texts. We then randomly select $500$ texts including the annotated name of the supplier, and ask the LLM to extract the name of the supplier.
    %The final set of texts has $98$ characters on average (min: $50$, max: $245$). For this task, we treat the initial model output as a self-explanation, as it can be interpreted as a span of important words towards the task.

    \item \textbf{Food hazard classification} on the \textit{Food Recall Incidents dataset}~\cite{Randl2024_FoodIncidents}. It contains the titles of official food recalls released by government and non-government organizations. We use expert annotations\footnote{Experts from \href{https://agroknow.com}{AgroKnow}}, identifying the specific reason for recalling the product, to extract matching spans in the texts. We then randomly selected $200$ texts and ask the model to classify the recall into one of the following classes:
    ``biological'' ($77$ texts),
    ``allergens'' ($53$),
    ``chemical'' ($29$),
    ``foreign bodies'' ($20$),
    ``organoleptic aspects'' ($12$), or
    ``fraud'' ($9$).
    The final set of texts has $95$ characters on average (min: $51$, max: $209$).

    \item \textbf{Sentiment classification} on an annotated version of the \textit{Movie Review Polarity dataset v2}~\cite{Pang2005_MovieReviewsDataset,Zaidan2008-movies}. We use the validation split which contains 200 labeled movie reviews (100 positive, 100 negative) from IMDB\footnote{https://www.imdb.com/} with human-annotated spans carrying high information towards the task. For each review, we extract a random one- to three-sentence-long snippet including at least one annotated important span, and ask the LLM to classify the sentiment of the snippet.
    We only use excerpts of the reviews to keep the task close to our first task, reduce the duration of the experiments, and increase the interpretability of texts and explanations for qualitative inspection.
    The final set of texts has an average length of $344$~characters (min: $48$, max: $986$) and up to six annotated spans.
\end{enumerate}
%Both tasks are designed to be easily interpretable by a human, given the short text length.
The main difference between the tasks is that the first task has a very confined set of important words (one to two per text), while the second task has the important words spread out over the whole sample.

\subsection{Evaluation metrics}
\label{sec:metrics}
As a preprocessing step, enabling the comparison of self-explanations to analytic explanations, we convert span-based explanations to saliency maps: for extractive self-explanations and human annotations, we find the first occurrence in the input text and compute the token indices. We then assign a saliency of $1$ to all tokens in these indices.
For counterfactual self-explanations, we find tokens of the original text that have been changed in the counterfactual and assign saliency of $1$.
Tokens not affected by the previous steps are assigned a saliency of $1\cdot10^{-9}$ to avoid zero division errors. Afterward, we normalize all saliency maps (LLM-generated, gradient, and attention-based) through division by their sum.
To assess the ability of an LLM to explain itself on the above tasks, we employ three quantitative evaluation metrics in addition to a qualitative assessment.

\subsubsection{Faithfullness:} A generally agreed upon measure of explanation quality is a faithfulness test by means of perturbing the model input~\cite{Abnar2020_AttentionFlow,Chefer2021_TransformerInterpretabilityBeyondAttentionVisualization,Liu2021_AttentionBasedExplanation}. In our case, we iteratively prompt the LLM after masking the input tokens from the most important to the least important (``\textit{high to low}'') and vice versa (``\textit{low to high}'') in steps of 
%$0.2 \cdot $total number of tokens. 
$0.2 \cdot |tokens|$.
If several tokens have the same importance, we randomly select one. For Gemma, we mask with the \texttt{<unk>} token. As Llama~3 does not have a pre-trained mask token, we mask with the token \texttt{\#\#\#}, which is often used for obscuring texts and should therefore be understood by the LLM.
Intuitively, for a faithful explanation method, the class predicted by the LLM should change very early in the \textit{high to low} test, as the removal of important tokens should alter the output, while it should change very late in the \textit{low to high} test, as the removal of unimportant tokens should not alter the LLM's assessment.

\vspace{-5pt}
\subsubsection{Text similarity:}
As there are multiple dimensions to measure the similarity between two texts, we employ several measures to assess it:
\textbf{(i)} Two simple complementing measures of token-count-based similarity are BLEU~\cite{Papineni2002_bleu} and ROUGE~\cite{Lin2004_rouge}. Both compare n-gram overlaps between the candidate and the reference translations. BLEU, which also adjusts for length discrepancies, emphasises Precision (normalises the overlap against the generated text) while ROUGE emphasises Recall (normalises against the reference text). As both of them ignore token/n-gram order we also apply the similarity ratio provided in Python's \texttt{difflib.SequenceMatcher}: this measure counts all matching tokens in the order of the reference and computes the ratio towards the \textit{average token count} of reference and candidate. 
\textbf{(ii)} To measure semantic similarity of a candidate towards a reference we employ BARTScore~\cite{Yuan2021_BARTScore}. Contrary to the previous metrics, this metric, based on the similarity of transformer embeddings, is defined on a logarithmic scale and therefore produces scores between negative infinity~(no semantic similarity) to zero~(high semantic similarity).

\vspace{-5pt}
\subsubsection{Similarity of saliency maps:}
In order to compare two explanations on feature importance level, we compute Pearson's~$r$ per (input) text as a measure of correlation. We prefer correlation over other metrics (such as accuracy) as correlation respects the order of tokens.
As normal distributions cannot be automatically assumed for these correlations, we present violin plots instead of average values.

\subsection{Results}

\begin{table}[t]
    \centering
    \caption{Performance of LLMs in entity extraction and sentiment classification. The min and max gradients are computed per input text per task, and averaged.}
    \label{tab:performance}
    \setlength{\tabcolsep}{4pt}
    \begin{tabular}{|c||c|cc||c|cc|}
        \hline

        & \multicolumn{3}{c||}{\textsc{Food}}
        & \multicolumn{3}{c|}{\textsc{Movies}}
        \\

        & \textbf{F$_1$}
        & \multicolumn{2}{c||}{\textbf{Gradient}}
        
        & \textbf{F$_1$}
        & \multicolumn{2}{c|}{\textbf{Gradient}}
        \\

        & \small macro
        & \small min
        & \small max
        
        & \small macro
        & \small min
        & \small max
        \\

        \hline
        \hline
    
        \textbf{Gemma-2B}
        & $0.30$
        & $-0.03$
        & $ 0.03$

        & $0.89$
        & $-0.02$
        & $ 0.02$
        \\

        \textbf{Gemma-7B}
        & $0.36$
        & $-0.28$
        & $ 0.21$

        & $0.90$
        & $-0.19$
        & $ 0.13$
        \\

        \textbf{Llama~3-8B}
        & $0.34$
        & $-1.05$
        & $ 1.04$

        & $0.95$
        & $-0.64$
        & $ 0.58$
        \\

        \hline
    \end{tabular}
\end{table}

\subsubsection{Food hazard classification:}
As shown in Table~\ref{tab:performance} all models show moderate, but above random, zero-shot performance on the first task.
At least for Gemma-7B and Llama~3-8B, we get gradient values completely different from 0, indicating that the gradients are not saturating and that gradient-based explanations are clearly distinguishable from noise.

\begin{figure}[ht]
    \flushleft

    {\tiny\hspace{35pt}%
    Correlation with human spans:%
    \hspace{100pt}%
    Correlation with extractive self-expl.:}

    \centering

    \rotatebox{90}{\hspace{10pt}\textbf{a)}~\textsc{Food}}
    \includegraphics[trim={20 35 10 10},clip,height=1.835cm]{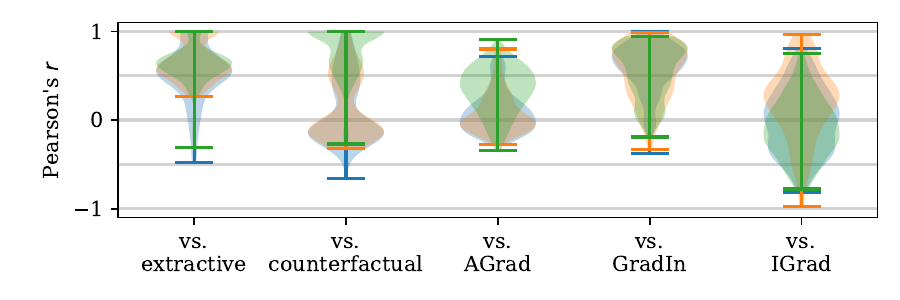}%
    \includegraphics[trim={203 35 10 10},clip,height=1.835cm]{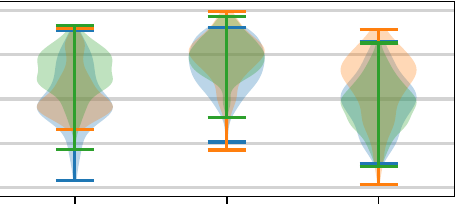}

    \rotatebox{90}{\hspace{17pt}\textbf{b)}~\textsc{Movies}}
    \includegraphics[trim={20 10 10 10},clip,height=2.3cm]{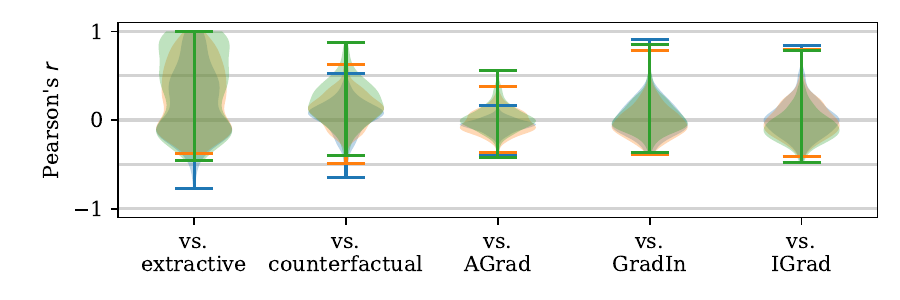}%
    \includegraphics[trim={203 10 10 10},clip,height=2.3cm]{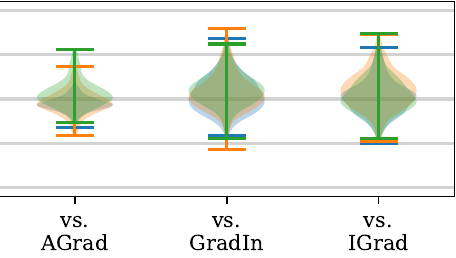}

    \flushleft

    \vspace{-15pt}

    {\tiny\hspace{46pt}\textbf{self-explanations}}

    \flushright

    \vspace{-10pt}

    \includegraphics[trim={40 0 40 0},clip,width=.66\linewidth]{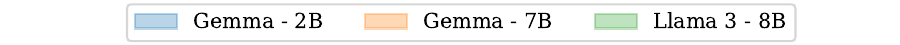}

    \caption{Per text Pearson's~$r$ correlation with human annotations (\textit{left}) and the LLM's extractive self-explanations (\textit{right}).}
    \label{fig:pearson}
\end{figure}

Figure~\ref{fig:pearson}~a) shows the per-text-correlation of the human annotations and the explainability methods. The figure shows a clear positive correlation of the human annotations with
extractive self-explanations and GradIn. Counterfactual self-explanations, as well as AGrad, are positively correlated for Llama~3 but uncorrelated for the Gemma models. IGrad shows no clear correlation for any of the models.
The correlations of extractive self-explanations and the analytic methods show a similar picture: GradIn in is positively correlated in all LLMs while AGrad only correlates for Llama~3 and IGrad show no clear trend.

\begin{figure}[ht]
    \centering

    \includegraphics[trim={0 0 0 30},clip,width=\linewidth]{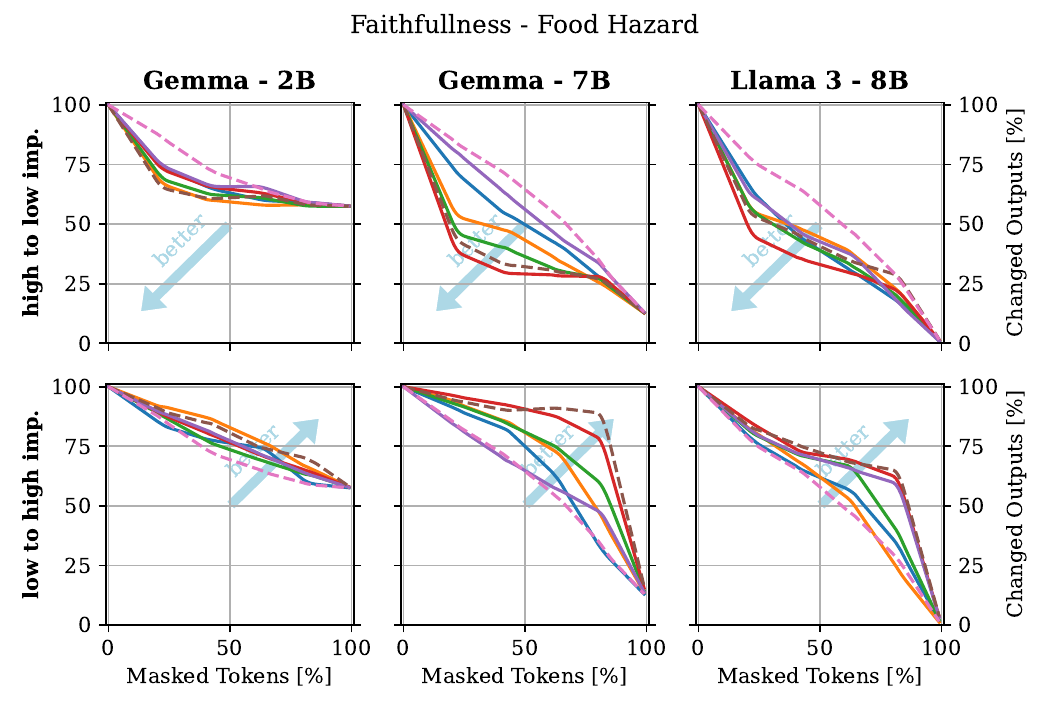}
    
    \includegraphics[trim={20 0 5 0},clip,width=\linewidth]{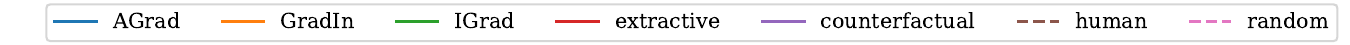}

    \caption{Faithfulness test for food hazard classification. Human-annotated spans (brown) help measure how easy it is to guess token importance from an external point of view.}
    \label{fig:faithfullness-food}
\end{figure}

To assess whether these correlations are linked to the faithfulness of the methods, we provide the perturbation curves (see Section~\ref{sec:metrics}) in Figure~\ref{fig:faithfullness-food}: while all curves show a clear early drop during ``high-to-low'' perturbation, proving that the explanations successfully indicate the most important tokens, only Llama~3 is able to retain model output when removing unimportant tokens (``low-to-high'').
Furthermore, the explanations of the two larger LLMs eventually change the model output for perturbations in both directions, while Gemma-2B’s explanations only change the output in up to 70\% of the cases (i.e., 0.3 or more, vertically). Further investigation shows, that Gemma-2B predicts all completely obscured texts to the most supported class ``biological''. This leads to not switching the label for all 30\% of samples that were originally predicted to this class. Gemma-7B classifies the majority ($92\%$) of occluded texts to the much less supported class ``foreign bodies''. Llama~3's output for completely occluded texts is more useful, stating a missing text in $83\%$ of cases. This different behavior of the two model families may be tied to the use of different occlusion tokens.

In general extractive self-explanations, human labels, and IGrad show the highest faithfulness over all models. For Llama~3 counterfactuals show high faithfulness while Gemma's counterfactuals are amongst the least faithful explanations.
Comparing faithfulness to Figure~\ref{fig:pearson}, we cannot find a clear connection.
%The span-based methods being the most successful ones in this task is intuitive, as for them, all tokens outside the span are assigned zero importance.

\begin{table}[h]
    \centering
    \caption{Counterfactual quality for both tasks. The reported similarity metrics are the mean over all (generated) texts that successfully change the model output.}
    \label{tab:counterfactuals}
    \resizebox{\textwidth}{!}{
    \setlength{\tabcolsep}{4pt}
    \begin{tabular}{|cc||c|cccc|c|}
        \hline

        &
        & \multirow{2}{*}{\textbf{Validity}}
        & \multirow{2}{*}{\textbf{Similarity}}
        & \multirow{2}{*}{\textbf{ROUGE-1}}
        & \multirow{2}{*}{\textbf{BLEU-1}}
        & \multirow{2}{*}{\textbf{ROUGE-L}}
        & \textbf{BART}
        \\

        &
        &
        & 
        & 
        & 
        & 
        & \textbf{Score}
        \\
    
        \hline
        \hline

%        \multirow{2}{*}{\rotatebox[origin=c]{90}{\textsc{Extr.~}}}
%        
%        & \textbf{Gemma-2B}
%        & $ 0.25$
%        & $ 0.79$
%        & $ 0.44$
%        & $ 0.48$
%        & $ 0.43$
%        & $-2.89$
%        \\
%        
%        & \textbf{Gemma-7B}
%        & $ 0.23$
%        & $ 0.92$
%        & $ 0.72$
%        & $ 0.80$
%        & $ 0.72$
%        & $-2.20$
%        \\
%        
%        & \textbf{Llama~3-8B}
%        & $ 0.44$
%        & $ 0.93$
%        & $ 0.83$
%        & $ 0.91$
%        & $ 0.80$
%        & $-2.36$
%        \\
%
%        \hline

        \multirow{2}{*}{\rotatebox[origin=c]{90}{\textsc{Food\hspace{4pt}}}}

        & \textbf{Gemma-2B}
        & $ 0.11$
        & $ 0.95$
        & $ 0.77$
        & $ 0.75$
        & $ 0.75$
        & $-2.37$
        \\
        
        & \textbf{Gemma-7B}
        & $ 0.41$
        & $ 0.94$
        & $ 0.79$
        & $ 0.70$
        & $ 0.73$
        & $-2.41$
        \\
        
        & \textbf{Llama~3-8B}
        & $ 0.29$
        & $ 0.97$
        & $ 0.83$
        & $ 0.89$
        & $ 0.86$
        & $-2.46$
        \\

        \hline

        \multirow{2}{*}{\rotatebox[origin=c]{90}{\textsc{Movie\hspace{2pt}}}}

        & \textbf{Gemma-2B}
        & $ 0.39$
        & $ 0.53$
        & $ 0.32$
        & $ 0.44$
        & $ 0.25$
        & $-3.17$
        \\
        
        & \textbf{Gemma-7B}
        & $ 0.95$
        & $ 0.67$
        & $ 0.52$
        & $ 0.61$
        & $ 0.49$
        & $-2.55$
        \\
        
        & \textbf{Llama~3-8B}
        & $ 0.94$
        & $ 0.85$
        & $ 0.81$
        & $ 0.83$
        & $ 0.81$
        & $-1.69$
        \\

        \hline
    \end{tabular}
    }
\end{table}

The low performance of the counterfactuals for the food hazard classification task with the Gemma models is also shown in the upper part of Table~\ref{tab:counterfactuals}: the generated explanations only change the LLM output in less than half of the samples. However, counterintuitively, we see a similarly bad validity in Llama~3. Further qualitative assessment of this artifact yields that all three models are able to identify important tokens, but don't consistently replace them with counterfactual evidence. The Gemma models rather highlight these with markdown (e.g. ``**salmonella**'' instead of ``salmonella''), while Llama replaces them with other hazards from the same class (e.g. ``e. coli'' instead of ``salmonella''). While in both cases the LLM fails to produce valid counterfactuals, these observations imply that more precise prompts, suggesting a class to change to, could improve validity. Additionally, as in Table~\ref{tab:counterfactuals}, similarity for valid counterfactuals is high.

\subsubsection{Sentiment classification:}
For our second task, we see much higher F$_1$-scores (see Table~\ref{tab:performance}) compared to the first task, increasing with model size.
Contrary, the average per-text-extrema of the gradients are much lower. This is expected, however, as in this task the importance is not focused on a few tokens, but spread out over the complete sample text. At least for Gemma-7B and Llama~3-8B, the values are large enough to exclude gradient saturation.

Contrary to the first task, the correlation plots in Figure~\ref{fig:pearson}~b) show a very clear picture: while the human-annotated spans are uncorrelated to all the other explainability methods, we see varying degrees of correlation with the extractive self-explanations, indicating that for some samples the LLM exactly reproduces the ground truth while for others it produces spans not overlapping at all. As in the first task, the correlation of extractive self-explanations with the analytic methods mimics the correlation of human annotations with the respective method, resulting in no correlation for all three methods.

%One interesting finding is that, for the analytic methods, the only method showing a negative correlation between the explanations for the output tokens ``Positive'' and ``Negative'' is IGrad~(
%Gemma-2B:  $\bar{r}=-0.24$,
%Gemma-7B:  $\bar{r}=-0.99$,
%Llama~3-8B: $\bar{r}=-0.20$
%). For AGrad~(
%Gemma-2B:  $\bar{r}=0.86$,
%Gemma-7B:  $\bar{r}=0.99$,
%Llama~3-8B: $\bar{r}=0.12$
%) and GradIn~(
%Gemma-2B:  $\bar{r}=0.56$,
%Gemma-7B:  $\bar{r}=0.99$,
%Llama~3-8B: $\bar{r}=0.27$
%) we see a positive correlation, indicating that syntactic information, which is independent of the label, has a high importance for both of them.

\begin{figure}[h]
    \centering

    \includegraphics[trim={0 0 0 30},clip,width=\linewidth]{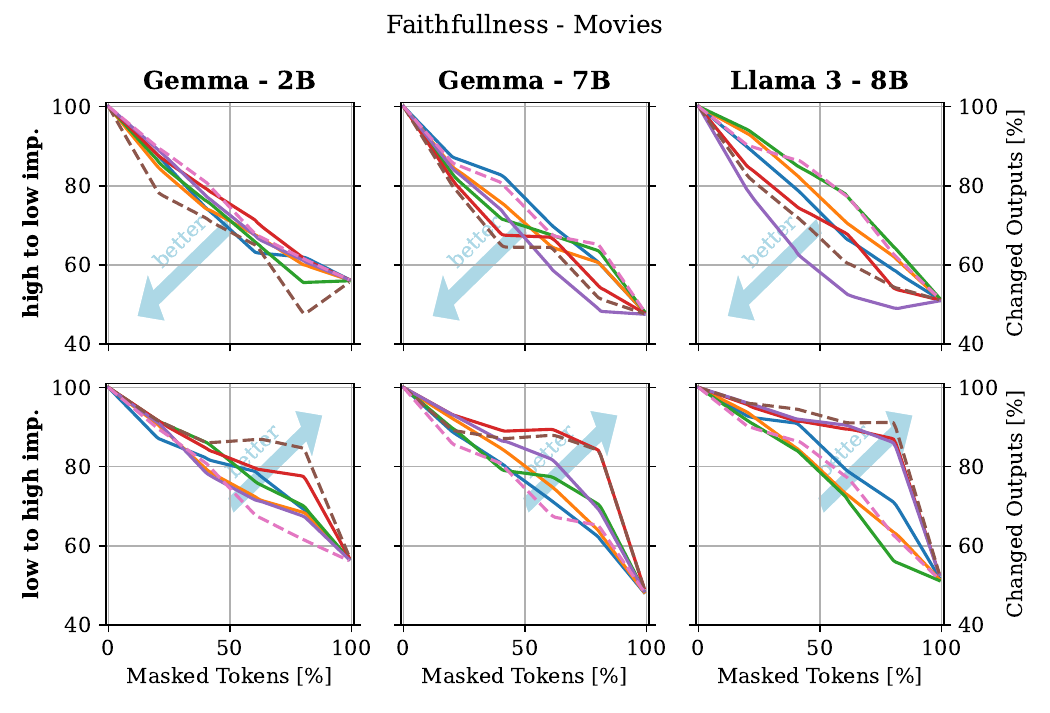}
    
    \includegraphics[trim={20 0 5 0},clip,width=\linewidth]{Faithfullness-Legend.pdf}

    \caption{Faithfullness test for the sentiment classification task. The human annotated spans (brown) provide a measure of how easy it is to guess token importance from an external point of view.}
    \label{fig:faithfullness-movies}
\end{figure}

The results of the faithfulness test, shown in Figure~\ref{fig:faithfullness-movies}, are much less clear than for the first task: For Gemma-7B and Llama~3-8B, the most faithful methods are counterfactual self-explanations and human explanations. The remaining methods perform comparably over all models and directions. The figure also shows that for this binary task, completely obscured texts only flip the label in $50\%$ of the cases: While Gemma-2B assigns all completely obscured samples to the ``negative'' class, there is no clear trend for the remaining two LLMs. Contrary to the first task, Llama~3 only states a missing review in $19\%$ of the samples.

The finding that, contrary to the first task, prompting for counterfactuals seems to work well for the sentiment classification task is reinforced by the lower part of Table~\ref{tab:counterfactuals}: For the two larger models we see successful flipping of the predicted class in at least 94\% of the cases. While counterfactuals overall achieve acceptable similarity, Llama~3-8B achieves the overall best semantic similarity for its counterfactuals.

\section{Discussion and Conclusion}
In all scenarios, self-explanations (incl. extractive) correlation with human annotations is higher on average compared to analytic explanations. Therefore, our answer to the research question \textbf{RQ1} is clearly yes.

Focusing on faithfulness (Figures \ref{fig:faithfullness-food}, \ref{fig:faithfullness-movies}), we show that the extractive self-explanations and the human-annotated ground truth perform better in the ``low to high'' test. This finding, however, does not necessarily mean higher faithfulness compared to analytic gradient/attention methods. Such methods often assign high importance to syntactically significant tokens (e.g., punctuation), crucial for language understanding but not necessarily important for the task. 
%Span-based explanations do not include this syntactic information.

Regarding \textbf{RQ2}, our presented results show that self-explanations are not generally correlated with the internal LLM dynamics. While we find correlations between extractive self-explanations and analytic explanations for objective tasks with clear token/task dependencies, such correlations are not guaranteed for subjective tasks that require reading between the lines. The different formats of LLM-generated explanations and analytic methods may partly account for this. Our food hazard classification task, however, shows that correlations are possible but not the rule.

Combining our answers to \textbf{RQ1} and \textbf{RQ2}, we argue that extractive self-explanations can be seen as the model's most probable explanations based on its training data. If there are correlations with analytic methods, they appear in tasks with a clear dependency between specific tokens and the supervision signal. In such cases, the correlation between self-explanations and analytic methods is indirect, with both being directly correlated with the ground truth but not each other. 
%Hence, they do not necessarily explain the inference process and need to be subject to additional quality checks in order to be useful.
Regardless of this finding, there is no advantage of analytic explanations over self-explanations in terms of faithfulness, as is known in literature~\cite{huang2023_CanLLMs}. As, human explanations 
%, delivered at a time preceding the model training 
are also found to be faithful, we argue that faithfulness of self-explanations stems from self-explanations mimicking human assessment.

\paragraph{Counterfactuals} We see that LLM-generated counterfactuals can produce highly faithful and informative explanations for the sentiment classification task. While these do not necessarily explain the LLM’s internal processes as previously discussed, they can provide information on whether the LLM correctly ``understood'' the assigned task. We further argue, that because counterfactuals can be easily validated by feeding them through the LLM, and our finding that valid counterfactuals were highly similar to the original in our experiments, they present an effective alternative to SHAP and LIME and an interesting field of further study.
We also find that prompt tuning for counterfactuals is essential for their success. Further research could examine provding specific classes to which the original text should be changed or redactions as suggested by \cite{madsen2024_selfexplanations}.

\paragraph{Limitations}
\textbf{(i)}~We chose to limit our methods for creating self-explanations to post-hoc and text-based approaches, as we consider them a natural way to use LLMs. Other approaches, for example creating the explanations at inference time like CoT~\cite{Wei2022_CoT}, or prompting for numerical word importance~\cite{huang2023_CanLLMs} exist but are not considered here.
\textbf{(ii)}~In this publication, we do not explore how different prompts affect the quality of the self-explanations tested in our experiments. We argue, that this is highly dependent on the LLM in question and hard to generalize. In our experiments we used a prompt we found working for all our tasks and models.
\textbf{(iii)}~Due to hardware limitations we only experiment on models up to 8B parameters. We are aware that larger models exist and can be accessed through APIs, but we need to be able to modify the LLMs source code in order to implement our attention- and gradient-based methods.

\begin{credits}
\subsubsection{\ackname} This work has been partially supported by project MIS 5154714 of the National Recovery and Resilience Plan Greece 2.0 funded by the European Union under the NextGenerationEU Program. Funding for this research has also been provided by the European Union’s Horizon Europe research and innovation programme EFRA (Grant Agreement Number 101093026). Funded by the European Union. Views and opinions expressed are however those of the author(s) only and do not necessarily reflect those of the European Union or European Commission-EU. Neither the European Union nor the granting authority can be held responsible for them. {\normalsize\euflag}

\subsubsection{\discintname} The authors declare no competing interests.
\end{credits}
%
% ---- Bibliography ----
%
% BibTeX users should specify bibliography style 'splncs04'.
% References will then be sorted and formatted in the correct style.
%
\bibliographystyle{splncs04}
\bibliography{sources}

\begin{thebibliography}{10}
\providecommand{\url}[1]{\texttt{#1}}
\providecommand{\urlprefix}{URL }
\providecommand{\doi}[1]{https://doi.org/#1}

\bibitem{Abnar2020_AttentionFlow}
Abnar, S., Zuidema, W.: Quantifying attention flow in transformers. arXiv preprint arXiv:2005.00928  (2020)

\bibitem{llama3modelcard}
AI@Meta: Llama 3 model card  (2024)

\bibitem{Almazrouei2023_Falcon}
Almazrouei, E., Alobeidli, H., Alshamsi, A., Cappelli, A., Cojocaru, R., Debbah, M., Étienne Goffinet, Hesslow, D., Launay, J., Malartic, Q., Mazzotta, D., Noune, B., Pannier, B., Penedo, G.: The falcon series of open language models (2023)

\bibitem{Bach2015_LRP}
Bach, S., Binder, A., Montavon, G., Klauschen, F., Müller, K.R., Samek, W.: On pixel-wise explanations for non-linear classifier decisions by layer-wise relevance propagation. PLoS ONE 10(7)  (2015)

\bibitem{Brown2020_FewShotLearning}
Brown, T., Mann, B., Ryder, N.e.a.: Language models are few-shot learners. In: Larochelle, H., Ranzato, M., Hadsell, R., Balcan, M., Lin, H. (eds.) Advances in Neural Information Processing Systems. vol.~33, pp. 1877--1901. Curran Associates, Inc. (2020)

\bibitem{Chefer2021_TransformerInterpretabilityBeyondAttentionVisualization}
Chefer, H., Gur, S., Wolf, L.: Transformer interpretability beyond attention visualization. In: 2021 IEEE/CVF Conference on Computer Vision and Pattern Recognition (CVPR). pp. 782--791 (2021)

\bibitem{Devlin2019_BERT}
Devlin, J., Chang, M.W., Lee, K., Toutanova, K.: Bert: Pre-training of deep bidirectional transformers for language understanding (2019)

\bibitem{Gurrapu2023_Rationalization}
Gurrapu, S., Kulkarni, A., Huang, L., Lourentzou, I., Batarseh, F.A.: Rationalization for explainable nlp: a survey. Frontiers in Artificial Intelligence  (sep 2023)

\bibitem{Hechtlinger2016_InputGradient}
Hechtlinger, Y.: Interpretation of prediction models using the input gradient (2016)

\bibitem{huang2023_CanLLMs}
Huang, S., Mamidanna, S., Jangam, S., Zhou, Y., Gilpin, L.H.: Can large language models explain themselves? a study of llm-generated self-explanations (2023)

\bibitem{Jiang2023_Mistral}
Jiang, A.Q., Sablayrolles, A., Mensch, A., Bamford, C., Chaplot, D.S., de~las Casas, D., Bressand, F., Lengyel, G., Lample, G., Saulnier, L., Lavaud, L.R., Lachaux, M.A., Stock, P., Scao, T.L., Lavril, T., Wang, T., Lacroix, T., Sayed, W.E.: Mistral 7b (2023)

\bibitem{Lewis2020_bart}
Lewis, M., Liu, Y., Goyal, N., Ghazvininejad, M., Mohamed, A., Levy, O., Stoyanov, V., Zettlemoyer, L.: {BART}: Denoising sequence-to-sequence pre-training for natural language generation, translation, and comprehension. In: Jurafsky, D., Chai, J., Schluter, N., Tetreault, J. (eds.) Proceedings of the 58th Annual Meeting of the Association for Computational Linguistics. pp. 7871--7880. Association for Computational Linguistics, Online (Jul 2020)

\bibitem{Li2017_occlusison}
Li, J., Monroe, W., Jurafsky, D.: Understanding neural networks through representation erasure (2017)

\bibitem{Li2023_UnderstandingICL}
Li, Z., Xu, P., Liu, F., Song, H.: Towards understanding in-context learning with contrastive demonstrations and saliency maps. CoRR  \textbf{abs/2307.05052} (2023)

\bibitem{Lin2004_rouge}
Lin, C.Y.: {ROUGE}: A package for automatic evaluation of summaries. In: Text Summarization Branches Out. pp. 74--81. Association for Computational Linguistics, Barcelona, Spain (Jul 2004)

\bibitem{Liu2021_AttentionBasedExplanation}
Liu, S., Le, F., Chakraborty, S., Abdelzaher, T.: On exploring attention-based explanation for transformer models in text classification. In: 2021 IEEE International Conference on Big Data (Big Data). pp. 1193--1203 (2021)

\bibitem{Liu_roberta}
Liu, Y., Ott, M., Goyal, N., Du, J., Joshi, M., Chen, D., Levy, O., Lewis, M., Zettlemoyer, L., Stoyanov, V.: Roberta: A robustly optimized bert pretraining approach (2019)

\bibitem{Lundberg2017_SHAP}
Lundberg, S.M., Lee, S.I.: A unified approach to interpreting model predictions. In: Guyon, I., Luxburg, U.V., Bengio, S., Wallach, H., Fergus, R., Vishwanathan, S., Garnett, R. (eds.) Advances in Neural Information Processing Systems 30, pp. 4765--4774. Curran Associates, Inc. (2017)

\bibitem{madsen2024_selfexplanations}
Madsen, A., Chandar, S., Reddy, S.: Are self-explanations from large language models faithful? (2024)

\bibitem{Montavon2017_DeepTaylor}
Montavon, G., Lapuschkin, S., Binder, A., Samek, W., Müller, K.R.: Explaining nonlinear classification decisions with deep taylor decomposition. Pattern Recognition  \textbf{65},  211–222 (May 2017), \url{http://dx.doi.org/10.1016/j.patcog.2016.11.008}

\bibitem{Pang2005_MovieReviewsDataset}
Pang, B., Lee, L.: Seeing stars: Exploiting class relationships for sentiment categorization with respect to rating scales. In: Proceedings of ACL. pp. 115--124 (2005)

\bibitem{Papineni2002_bleu}
Papineni, K., Roukos, S., Ward, T., Zhu, W.J.: {B}leu: a method for automatic evaluation of machine translation. In: Isabelle, P., Charniak, E., Lin, D. (eds.) Proceedings of the 40th Annual Meeting of the Association for Computational Linguistics. pp. 311--318. Association for Computational Linguistics, Philadelphia, Pennsylvania, USA (Jul 2002)

\bibitem{Raffel2020_T5}
Raffel, C., Shazeer, N., Roberts, A., Lee, K., Narang, S., Matena, M., Zhou, Y., Li, W., Liu, P.J.: Exploring the limits of transfer learning with a unified text-to-text transformer. J. Mach. Learn. Res.  \textbf{21}(1) (jan 2020)

\bibitem{Randl2024_FoodIncidents}
Randl, K., Karvounis, M., Marinos, G., Pavlopoulos, J., Lindgren, T., Henriksson, A.: Food recall incidents (Mar 2024)

\bibitem{Ribeiro2016_LIME}
Ribeiro, M.T., Singh, S., Guestrin, C.: "why should i trust you?": Explaining the predictions of any classifier. p. 1135–1144. KDD '16, Association for Computing Machinery, New York, NY, USA (2016)

\bibitem{Shrikumar2017_GradIn}
Shrikumar, A., Greenside, P., Kundaje, A.: Learning important features through propagating activation differences. In: Proceedings of the 34th International Conference on Machine Learning - Volume 70. p. 3145–3153. ICML'17, JMLR.org (2017)

\bibitem{Socher2013_SST}
Socher, R., Perelygin, A., Wu, J., Chuang, J., Manning, C.D., Ng, A., Potts, C.: Recursive deep models for semantic compositionality over a sentiment treebank. In: Yarowsky, D., Baldwin, T., Korhonen, A., Livescu, K., Bethard, S. (eds.) Proceedings of the 2013 Conference on Empirical Methods in Natural Language Processing. pp. 1631--1642. Association for Computational Linguistics, Seattle, Washington, USA (Oct 2013)

\bibitem{Sundarajan2017_IntegratedGradients}
Sundararajan, M., Taly, A., Yan, Q.: Axiomatic attribution for deep networks. In: Proceedings of the 34th International Conference on Machine Learning - Volume 70. p. 3319–3328. ICML'17, JMLR.org (2017)

\bibitem{gemmateam2024_gemma}
Team, G., Mesnard, T., Hardin, C., et~al.: Gemma: Open models based on gemini research and technology (2024)

\bibitem{Touvron2023_Llama2}
Touvron, H., Martin, L., Stone, K., et~al.: Llama 2: Open foundation and fine-tuned chat models (2023)

\bibitem{Vaswani2017_Transformer}
Vaswani, A., Shazeer, N., Parmar, N., Uszkoreit, J., Jones, L., Gomez, A.N., Kaiser, L., Polosukhin, I.: Attention is all you need (2017)

\bibitem{Verma2022_counterfactual}
Verma, S., Boonsanong, V., Hoang, M., Hines, K.E., Dickerson, J.P., Shah, C.: Counterfactual explanations and algorithmic recourses for machine learning: A review (2022)

\bibitem{Wei2022_CoT}
Wei, J., Wang, X., Schuurmans, D., Bosma, M., Ichter, B., Xia, F., Chi, E.H., Le, Q.V., Zhou, D.: Chain-of-thought prompting elicits reasoning in large language models. In: Koyejo, S., Mohamed, S., Agarwal, A., Belgrave, D., Cho, K., Oh, A. (eds.) Advances in Neural Information Processing Systems 35: Annual Conference on Neural Information Processing Systems 2022, NeurIPS 2022, New Orleans, LA, USA, November 28 - December 9, 2022 (2022)

\bibitem{Yuan2021_BARTScore}
Yuan, W., Neubig, G., Liu, P.: Bartscore: Evaluating generated text as text generation. In: Ranzato, M., Beygelzimer, A., Dauphin, Y., Liang, P., Vaughan, J.W. (eds.) Advances in Neural Information Processing Systems. vol.~34, pp. 27263--27277. Curran Associates, Inc. (2021)

\bibitem{Zaidan2008-movies}
Zaidan, O., Eisner, J.: Modeling annotators: {A} generative approach to learning from annotator rationales. In: Lapata, M., Ng, H.T. (eds.) Proceedings of the 2008 Conference on Empirical Methods in Natural Language Processing. pp. 31--40. Association for Computational Linguistics, Honolulu, Hawaii (Oct 2008), \url{https://aclanthology.org/D08-1004}

\bibitem{Zhao2024_XLLM}
Zhao, H., Chen, H., Yang, F., Liu, N., Deng, H., Cai, H., Wang, S., Yin, D., Du, M.: Explainability for large language models: A survey  \textbf{15}(2) (feb 2024)

\end{thebibliography}
\end{document}